\documentclass[journal]{IEEEtran}
\usepackage{cite}
\usepackage{times}
\usepackage{epsfig}
\usepackage{graphicx}
\usepackage{amsmath}
\usepackage{amssymb}
\usepackage{caption}
\usepackage{graphicx, subfig}
\usepackage{amssymb}
\usepackage{array}
\usepackage{booktabs}
\usepackage{multirow}
\usepackage{color}
\begin{document}
\title{RUArt: A Novel Text-Centered Solution for Text-Based Visual Question Answering}

\author{Zan-Xia Jin, Heran Wu, Chun Yang, Fang Zhou, Jingyan Qin, Lei Xiao, Xu-Cheng Yin,~\IEEEmembership{Senior Member,~IEEE}% <-this % stops a space
\thanks{ Zan-Xia Jin and Heran Wu contributed equally to this work. Corresponding author:
Xu-Cheng Yin.}
\thanks{Zan-Xia Jin, Heran Wu, Chun Yang, and Fang Zhou are with the Department of Computer Science and Technology, School of Computer and Communication Engineering, University of Science and Technology Beijing, Beijing 100083, China (e-mail: zanxiajin@xs.ustb.edu.cn; heranwu@yeah.net; chunyang@ustb.edu.cn; zhoufang@ies.ustb.edu.cn).}
\thanks{Jingyan Qin is with the Department of Computer Science and Technology,
School of Computer and Communication Engineering, University of Science
and Technology Beijing, Beijing 100083, China, also with the Department of Industrial Design, School of Mechanical Engineering, University of Science and Technology Beijing, Beijing
100083, China (e-mail: qinjingyanking@foxmail.com)}
\thanks{Lei Xiao are with Tencent Technology (Shenzhen) Company Limited, Shenzhen 518057, China (e-mail: shawnxiao@tencent.com).}
\thanks{Xu-Cheng Yin is with the Department of Computer Science and Technology,
School of Computer and Communication Engineering, University of Science
and Technology Beijing, Beijing 100083, China, also with the Institute of
Artificial Intelligence, University of Science and Technology Beijing, Beijing
100083, China, and also with the USTB-EEasyTech Joint Laboratory of
Artificial Intelligence, University of Science and Technology Beijing, Beijing
100083, China (e-mail: xuchengyin@ustb.edu.cn)}% <-this % stops a space
}

\maketitle

\begin{abstract}
Text-based visual question answering (VQA) requires to read and understand text in an image to correctly answer a given question. However, most current methods simply add optical character recognition (OCR) tokens extracted from the image into the VQA model without considering contextual information of OCR tokens and mining the relationships between OCR tokens and scene objects. In this paper, we propose a novel text-centered method called RUArt (Reading, Understanding and Answering the Related Text) for text-based VQA. Taking an image and a question as input, RUArt first reads the image and obtains text and scene objects. Then, it understands the question, OCRed text and objects in the context of the scene, and further mines the relationships among them. Finally, it answers the related text for the given question through text semantic matching and reasoning. We evaluate our RUArt on two text-based VQA benchmarks (ST-VQA and TextVQA) and conduct extensive ablation studies for exploring the reasons behind RUArt's effectiveness. Experimental results demonstrate that our method can effectively explore the contextual information of the text and mine the stable relationships between the text and objects.
\end{abstract}

\begin{IEEEkeywords}
Attention mechanism, computer vision, machine reading comprehension, natural language processing, visual question answering.
\end{IEEEkeywords}

% \fi
\IEEEpeerreviewmaketitle

\section{Introduction}
\IEEEPARstart{V}{isual} question answering (VQA) is a comprehensive problem involving natural language processing and computer vision. It requires to analyze both the natural language question and the image visual content simultaneously and answer the question about the image \cite{li2019visual}. In recent years, VQA has witnessed a lot of success \cite{yu2019deep,anderson2018bottom,ben2017mutan,li2019relation,GaoJYLHWL19}. Text-based VQA \cite{biten2019scene,singh2019towards,KembhaviSSCFH17,kafle2018dvqa} is a specific type of VQA, which needs reading and understanding the textual content in an image and answering the given question. However, the current VQA models fail catastrophically on text-based questions\cite{biten2019scene,singh2019towards}!

\begin{figure}
  \centering
  \setlength{\abovecaptionskip}{0.2cm}
  \includegraphics[width=0.49\textwidth]{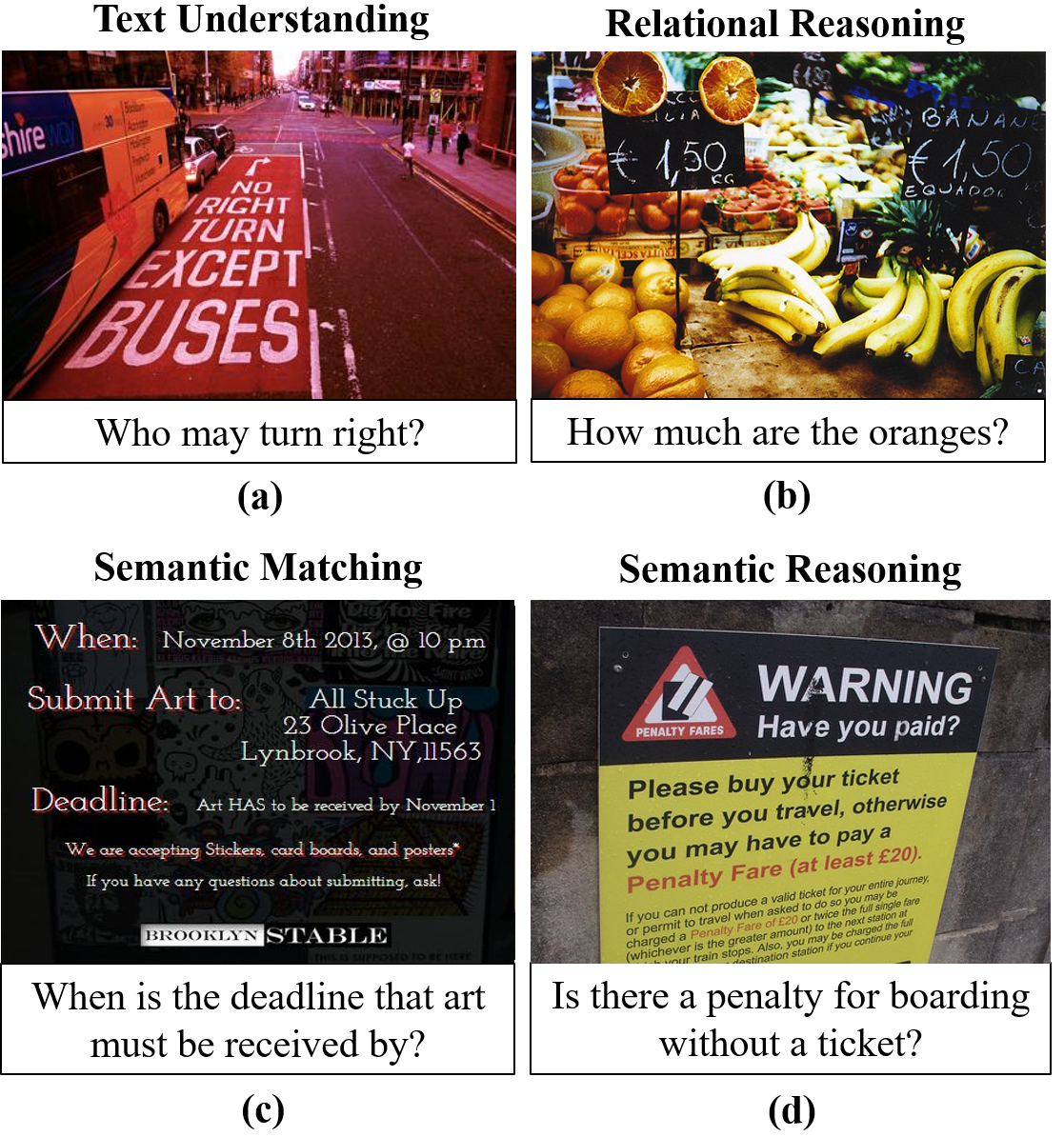}
  \caption{Text understanding, relation reasoning, semantic matching and reasoning are critical to answer the given question in text-based VQA. Examples are from ST-VQA \cite{biten2019scene} and TextVQA \cite{singh2019towards}.}
  \label{TVQA}
  \vspace{-0.6cm}
\end{figure}

The generic VQA \cite{antol2015vqa} mostly focuses on the questions about recognition of objects, attributes and activities in the image. And the text-based VQA is a little different that asks questions about the text embedded in the image, as displayed in Fig. \ref{TVQA}. Therefore, the key points of previous VQA models are mostly related to the localization of relevant object regions and the fine-grained object recognition, while the key to solving text-based VQA task is text recognition and machine reading comprehension. This is one of the reasons why text-based VQA task cannot be solved by existing VQA models, which cannot effectively read and utilize textual content in the image \cite{singh2019towards}. Therefore, text-based VQA is still a very challenging task, and requires more attention to the reading comprehension of the OCRed text.

For scene text-based VQA, two competitions ST-VQA\footnote{https://rrc.cvc.uab.es/?ch=11} \cite{biten2019scene} and TextVQA\footnote{https://textvqa.org/} \cite{singh2019towards} were put forward recently. The VTA  method, the winner of ST-VQA \cite{biten2019icdar},  encodes the question and the text with BERT \cite{devlin2018bert}, and decodes the answer with a similar model of the Bottom-Up and Top-Down strategy \cite{anderson2018bottom}. LoRRA \cite{singh2019towards} adopts the same architecture of the VQA components for getting fused OCR-question features and image-question features, and concatenates these two fused features to train a classifier. However, these methods simply add OCR tokens into the existing VQA models as the separate input, and do not make use of the relationships between OCR tokens. More recently, Gao et al. \cite{Gao_2020_CVPR} proposed a multi-modal graph neural networks (MM-GNN) to represent three modalities (i.e., visual, semantic, and numeric) in an image, and three aggregators to guide the flow of information between the various modalities. Hu et al. \cite{Hu_2020_CVPR} introduced a multi-modal transformer architecture (M4C), which fuses different modalities homogeneously by doing self-attention to model inter- and intra- modality context. These methods begin to realize the importance of the relationships between OCR tokens, but they still treat the OCR tokens as independent objects, rather than ordered words in a sentence which can convey more rich semantic information. In addition, although they explore the relationship between OCR features and object visual features, they ignore the natural semantic information of object properties (i.e., names and attributes). Moreover, the majority of these methods still solve the text-based VQA as a classification task, rather than a semantic matching and reasoning task. Specifically, the current models for text-based VQA have the following three limitations:

1) They pay less attention to the text reading comprehension. In many cases, the text embedded in the image may be one or more sentences, it is necessary to understand the context of OCR tokens to answer the given question. For example, to answer the question ``\emph{Who may turn right?}"  in Fig. \ref{TVQA}(a), the model needs to read and comprehend the whole OCRed text ``\emph{No right turn except buses}" within the image.

2) They ignore the relationships between the text and the corresponding object properties. In reality, the majority of images contain multiple objects and texts, such as the example in Fig. \ref{TVQA}(b). And the object properties (i.e., names and attributes) contain rich semantic information, which is in the same semantic space as OCRed text and questions, so it is reliable to establish the relationships between them.

3) They normally select answers from a fixed answer pool by training a classifier, and lack text semantic matching and reasoning in answer prediction. However, to answer the text-based questions, the model requires capturing the semantic association between the question and the answer candidates, or even reasoning based on the relevant OCR tokens, such as the examples in Fig. \ref{TVQA}(c) and \ref{TVQA}(d).

To deal with the above issues and answer the question with the true text, we propose a novel text-centered solution named RUArt (Reading, Understanding and Answering the Related Text) for text-based VQA. Taking an image and a question as input, RUArt first reads the image and obtains text and scene objects. Then, it understands the question and OCRed text by reading the context of the text, and makes relational reasoning between the text and the object based on semantics and position. Finally, it answers the related text for the given question through text semantic matching and reasoning. In summary, the main contributions of this work are three-fold:

\begin{itemize}
\item To exactly understand the text embedded in the image, we construct the OCR context according to the natural reading order of the text, and fully explore the contextual information of OCR tokens via a machine reading comprehension model.  Here, SDNet \cite{zhu2018sdnet} is used as a typical technique for machine reading comprehension.
\item To really capture the dependence between the text and  its corresponding objects, we mine the relationships between the OCR tokens and the objects in a scene by conducting semantic and positional attentions.
\item To fully utilize a variety of semantic information (question, text, and object) in answer prediction, we unify the multi-modal input into the context-dependent text, and predict the answer through text semantic matching and reasoning.
\end{itemize}

The rest of the paper is organized as follows: Section II summarizes the related work. Section III elaborates our work. In Section IV, we demonstrate experimental results on several datasets. Finally, we conclude our work in Section V.
%-------------------------------------------------------------------------
\section{Related Work}
%-------------------------------------------------------------------------
\subsection{Visual Question Answering}
Visual question answering is to provide an accurate natural language answer by understanding the given image and the question. Since one early VQA dataset was released in 2015 \cite{antol2015vqa}, VQA has attracted a large number of researchers from the natural language processing and computer vision communities. There have been a lot of successes in VQA in recent years. Yang et al. \cite{yang2016stacked} proposed a stacked attention network to learn the attention iteratively. Fukui et al. \cite{fukui2016multimodal}, Kim et al. \cite{kim2016hadamard}, Yu et al. \cite{yu2017multi} and Ben et al. \cite{ben2017mutan} exploited different multimodal bilinear pooling methods that integrate the visual features from the image with the textual features from the questions to predict the attention \cite{yu2019deep}. Anderson et al. \cite{anderson2018bottom} introduced a combined bottom-up and top-down attention mechanism that enables attention to be calculated at the level of objects and other salient image regions. Liang et al. \cite{Liang0CLH18} proposed a focal visual-text attention model for sequential data, which makes use of a hierarchical process to dynamically determine what media and what time focused on in the sequential data to answer the question. Nevertheless, how to model the complex interactions between these two different modalities is not an easy work. Li et al. \cite{li2019visual} represented the image content explicitly by the natural language using the image captioning method. Almost all VQA algorithms pose it as a classification problem in which each class is synonymous with a particular answer \cite{kafle2018dvqa}. In our work, we consider text-based VQA task as a semantic matching problem rather than a classification problem.
%-------------------------------------------------------------------------
\subsection{Text-Based Visual Question Answering}
Text-based VQA requires reading and understanding the textual information in an image that could have an correct answer towards the given question. For scene text-based VQA, two competitions ST-VQA \cite{biten2019scene} and TextVQA \cite{singh2019towards} are put forward recently. According to the published technical reports of these two competitions, the majority of the current approaches integrate the OCRed text into the existing VQA model to solve this new problem. LoRRA \cite{singh2019towards} uses the same architecture of the VQA components to get the combined OCR-question features and image-question features. The winning team of TextVQA 2019 followed the framework of LoRRA and applied multimodal factorized high-order pooling \cite{yu2018beyond} for multimodal fusion. For the ST-VQA challenge \cite{biten2019icdar}, the VTA method proposed by the winner is similar to the Bottom-Up and Top-Down method \cite{anderson2018bottom} with the BERT \cite{devlin2018bert} to encode the question and text. Different from the above methods, the QAQ \cite{biten2019icdar} method uses an unified end-to-end trainable oriented text spotting network for simultaneous detection and recognition. The Clova AI OCR \cite{biten2019icdar} method adopts MAC network \cite{hudson2018compositional} for combining visual cues and questions embedded with BERT \cite{devlin2018bert}, and uses pointer network for pointing coordinates of text boxes that match answers. More recently, Gao et al. \cite{Gao_2020_CVPR} proposed a multi-modal graph neural networks to represent three modalities (i.e., visual, semantic, and numeric) in an image, and three aggregators to guide the flow of information between the various modalities. Hu et al. \cite{Hu_2020_CVPR} introduced a multi-modal transformer architecture (M4C), which fuses different modalities homogeneously by doing self-attention to model inter- and intra- modality context. However, modeling the complex interactions between different modalities is not an easy work \cite{li2019visual}. In contrast to struggling on multimodal feature fusion, in our work, we unify all the input information by the plain text so as to convert text-based VQA into a text-only question answering (QA) problem, and a lot of QA \cite{chen2017reading,jin2019health,Jin2020Ranking,zhu2018sdnet} techniques are available and can be used.

\subsection{Object and OCR Relation Modeling}
Relations between objects in an image have been explored in many works for high-level applications, such as image retrieval, image captioning and VQA.  Chen et al. \cite{chen2012understanding} proposed an object relation network, a graph model representing the most probable meaning of the objects and their relations in an image. Hu et al. \cite{hu2018relation} dealt with a set of objects simultaneously through interactions between their appearance features and geometries, thus modeling their relations. The semantics induced learner module \cite{zhou2018object}, subtly incorporates semantics dependencies into the predicate classification model to predict object relations in one-shot manner. Li et al. \cite{li2019relation} proposed a graph-based relation encoder to learn both explicit and implicit relations between visual objects via graph attention networks. Song et al. \cite{song2019image} investigated object-to-object relations for scene recognition, including co-occurring frequency and sequential representation of object-to-object relations. CRA-Net, proposed by \cite{Peng0WWH19}, devises two question-adaptive relation attention modules that can extract not only the fine-grained and binary relations but also the more sophisticated trinary relations. Han et al. \cite{HanSLYS18} proposed a VSA-Net to detect relations in the image and designed a novel SO-layer to distinguish between the subject and the object. Jin et al. \cite{JinZG00Z19} took into account the object relations in video question answering task, in order to capture motions and other potential relations among the objects. To the best of our knowledge, there are few studies on the relations between OCRed text and objects in the image. Singh et al. \cite{Singh0SC19} integrated visual cues, textual cues and rich knowledge bases, and performed reasoning using a gated graph neural networks. However, it only learns relevant facts from prior knowledge, but ignores relations that may only exist in the given image, such as the price of a product. This kind of relationship is sometimes the information we want to obtain more when we look at the picture, which is one of the research focuses of this paper.
%------------------------------------------------------------------------
\section{Approach}
\begin{figure*}
  \centering
  \includegraphics[width=0.99\textwidth]{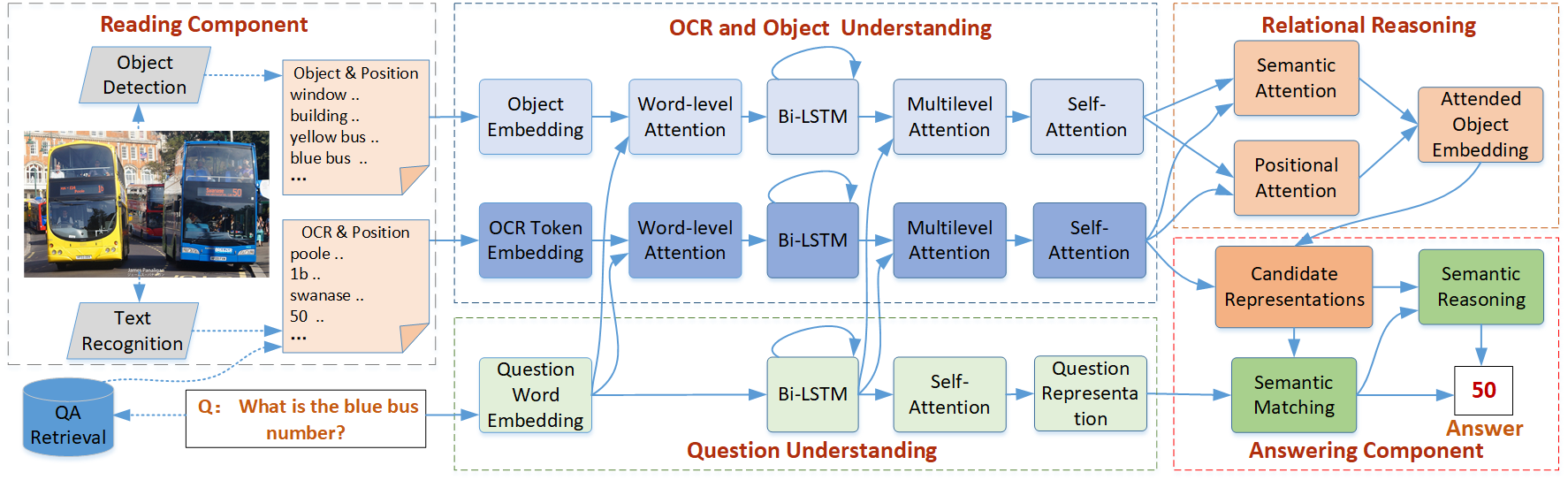}
  \caption{Overview of RUArt (Reading, Understanding and Answering the Related Text). Taking an image and a question as input, RUArt reads the image to obtain the text and objects, understands the question, OCRed text and objects in the context of the scene, and answers the related text for the given question. Dashed lines indicate that the text recognition module, object detection module and QA retrieval module are not jointly trained.}
  \label{framework}
  \vspace{-0.3cm}
\end{figure*}
\subsection{Overview of RUArt}
In this section, we introduce the architecture of our novel model RUArt for text-based VQA. At a high level, our model contains three components: (i) a \textbf{reading component} to extract the OCR tokens and objects from the image via the pre-trained OCR models and object detection models; (ii) an \textbf{understanding component} to understand the information conveyed by the question, OCRed text and objects. This component is divided into three modules: question understanding for comprehending the meaning of the question (\emph{Sec. B}), OCR and object understanding for exploring the contextual information of the OCR tokens and the objects via a machine reading comprehension model (\emph{Sec. C}), and relational reasoning for finding the relationships between the OCR tokens and the objects (\emph{Sec. D}); (iii) an \textbf{answering component} to answer questions by selecting responses from the OCRed text or additional text (\emph{Sec. E}). The overall model is shown in Fig. \ref{framework}. Note that the text recognition module and the object detection module can be any OCR model and any object detection model, and they are not jointly trained in our experiments.

\subsection{Question Understanding}
Given a question, question understanding module serves to comprehend the meaning of the sentence and produce the representation of the question. In order to generate a high level of understanding of the question, the multi-layer BiLSTMs \cite{Graves1997Long} is used and followed by a self-attention layer \cite{vaswani2017attention}.

In detail, we first encode each word of the question with 300-dim GloVe \cite{Pennington2014Glove} embedding and 768-dim BERT-Base \cite{devlin2018bert} contextual embedding similar to \cite{zhu2018sdnet}. Then we adopt the three-layer BiLSTMs to capture contextual information within question. Suppose the word embeddings of the question are $\{w_{1}^{Q}, \ldots, w_{q}^{Q}\}$, we can obtain a question representation $\{h_{1}^{Q}, \ldots, h_{q}^{Q}\} = \operatorname{BiLSTM}(\{w_{1}^{Q}, \ldots, w_{q}^{Q}\})$.
Next, we conduct self-attention on the question to extract relationships between words and capture important concepts of the question. The attended vectors from the question to itself are $\{\hat{h}_{1}^{Q}, \ldots, \hat{h}_{q}^{Q}\}$, defined as,
\begin{equation}
\begin{array}{c} {\alpha_{i j} = \operatorname{softmax} (\operatorname{ReLU}(U h_{i}^{Q}) D \operatorname{ReLU}(U h_{j}^{Q}))}, \\ {\hat{h}_{i}^{Q}=\sum_{j} \alpha_{i j} h_{j}^{Q}},\end{array}
\end{equation}
where $D \in \mathbb{R}^{k \times k}$ is a diagonal matrix and $U \in \mathbb{R}^{d \times k}$, $k$ is the attention hidden size \cite{HuangZSC18}. Finally, we condense the question representation into one vector, $u^{Q}=\sum_{i} \beta_{i} \hat{h}_{i}^{Q},$ where $\beta_{i} \propto \exp ({w}^{T} \hat{h}_{i}^{Q})$ and ${w}$ is a parameterized vector..

Similar to \cite{zhu2018sdnet}, to simplify notation, we define the attention function above as $\operatorname{Attn}(A,B,C)$, which means we compute the attention weight $\alpha_{i j}$ based on two sets of vectors $A$ and $B$, and use the weight to linearly combine vector set C. Therefore, the self-attention above can be simplified as $\hat{{h}}_{i}^{Q}=\operatorname{Attn}({h}_{i}^{Q},\{{h}_{j}^{Q}\}_{j=1}^{q},\{{h}_{j}^{Q}\}_{j=1}^{q})$.

\subsection{OCR and Object Understanding}
We use the pre-trained OCR model and object detection model to extract the OCR tokens and objects from images, respectively. We can also add relevant answers retrieved by the QA system as additional text. As mentioned above, the majority of existing text-based VQA methods cannot fully understand the contextual information of the OCR tokens. In this study, we first construct an OCR context with OCR tokens and tokens' positions according to natural reading order (i.e., from left to right and top to bottom), and then explore the contextual information of OCR tokens via a machine reading comprehension (MRC) model. By using the MRC method, we can align the question and OCR context in order to locate on the most relevant OCR tokens regarding the question. At the same time, objects are represented by the word embeddings of their corresponding names and attributes, and are processed the same way as OCR tokens in this module. Note that the understanding process of OCR tokens and objects can adopt any MRC model, and SDNet \cite{zhu2018sdnet} is used here as a typical technique. SDNet consists of the word-level attention layer, BiLSTMs layer, multilevel attention layer, and self-attention layer.

\textbf{Word-level Attention. } For providing the direct word information in the question to the context, we conduct word-level attention from question to context (OCR tokens or objects). We first encode each word of the OCR text and objects with 300-dim fastText \cite{GraveMJB17} embedding, which can generate word embeddings even for OOV tokens, and 768-dim BERT-Base \cite{devlin2018bert} contextual embedding.  Suppose the word embeddings of context are $\{{w}_{1}^{C}, \ldots, {w}_{m}^{C}\} \subset \mathbb{R}^{d}$, and then the word-level attention is  $\hat{{w}}_{i}^{C}=\operatorname{Attn}({w}_{i}^{C},\{{w}_{j}^{Q}\}_{j=1}^{q},\{{w}_{j}^{Q}\}_{j=1}^{q})$, where $\operatorname{Attn}()$ is defined above.
Thereafter, the attended vectors from question to context are $\{\hat{w}_{1}^{C}, \ldots, \hat{w}_{m}^{C}\}$.

\textbf{Multilevel Attention. } After word-level attention, we use two separate BiLSTMs to form contextual understanding for question and context (OCR tokens or objects). There are
\begin{equation}
\begin{array}{c}
{h_{1}^{C, k}, \ldots, h_{m}^{C, k}=\operatorname{BiLSTM}(h_{1}^{C, k-1}, \ldots, h_{m}^{C, k-1})}, \\
{h_{1}^{Q, k}, \ldots, h_{q}^{Q, k}=\operatorname{BiLSTM}(h_{1}^{Q, k-1}, \ldots, h_{q}^{Q, k-1})}, \\
{h_{i}^{C, 0}=[f_{i}^{C};b_{i}^{C};\hat{{w}}_{i}^{C} ; f_{w_{i}^{C}}]},
~~{h_{j}^{Q, 0}=[g_{j}^{Q};b_{j}^{Q}]},
\end{array}
\end{equation}
where $1 \leq k \leq K$ and $K$ is the number of BiLSTMs layers. $f_{i}^{C},g_{j}^{Q}$, $b_{i}^{C}$ and $b_{j}^{Q}$ are the fastText word embedding, GloVe word embedding and BERT contextual embedding, respectively. $f_{w_{i}^{C}}$ is a feature vector, including 12-dim POS embedding and 8-dim NER embedding \cite{chen2017reading}.

Multiple layers of BiLSTMs can extract different levels of understanding of each word. An approach that utilizes all the information from the word embedding level up to the highest level representation would be substantially beneficial for understanding both the question and the context, which has been proven many times in MRC \cite{HuangZSC18,zhu2018sdnet}. In order to fuse all levels of information in the question to the context (OCR tokens or objects), we follow the approach in FusionNet \cite{HuangZSC18} and conduct $K + 1$ times of multilevel attention from question to context, defined as,
\begin{equation}
\begin{aligned}
m_{i}^{(k), C}=\operatorname{Attn}(\mathrm{HoW}_{i}^{C},\{\mathrm{HoW}_{j}^{Q}\}_{j=1}^{q},
\{h_{j}^{Q, k}\}_{j=1}^{q}),
\end{aligned}
\end{equation}
where $1 \leq k \leq K+1$ and history-of-word vectors are
$\mathrm{HoW}_{i}^{C}=[f_{i}^{C};b_{i}^{C};h_{i}^{C, 1};\ldots; h_{i}^{C, k-1}]$,
$\mathrm{HoW}_{j}^{Q}=[g_{j}^{Q};b_{j}^{Q};h_{j}^{Q, 1};\ldots; h_{j}^{Q, k-1}].$

Similar to the question understanding, we conduct self-attention on OCR tokens and objects to establish direct correlations between all pairs of words, respectively. An additional RNN layer is applied to obtain the contextual representation for OCR tokens and objects.

\subsection{Relational Reasoning}
The output representations of OCR and object understanding module already contain rich information about the OCR tokens, objects and the question. However, the correlations between OCR tokens and objects have not been found, including semantic and positional relationships. To model these two relations between them, we conduct semantic attention and positional attention from objects to OCR tokens to obtain attended features $\hat{u}_{i}^{S},~\hat{u}_{i}^{P}$ respectively.

\textbf{Semantic Attention. } We conduct semantic attention from objects to OCR tokens based on semantic embeddings of them, which are the output of the OCR and object understanding module. Suppose the semantic embeddings of the OCR tokens and objects are $\{u_{1}^{O}, \ldots, u_{o}^{O}\} \subset \mathbb{R}^{d}$ and $\{u_{1}^{D}, \ldots, u_{n}^{D}\} \subset \mathbb{R}^{d}$, respectively, the attended feature is obtained as:
\begin{equation}
\hat{u}_{i}^{S}=\operatorname{Attn}(u_{i}^{O},\{u_{j}^{D}\}_{j=1}^{n},\{u_{j}^{D}\}_{j=1}^{n}).
\end{equation}

\textbf{Positional Attention. } We conduct positional attention from objects to OCR tokens based on positional embeddings and semantic embeddings of them. The positional embedding is a 8-dimensional location feature based on the OCR token's relative bounding box coordinates, which is defined as  $[x_1/W_{\operatorname{im}}, y_1/H_{\operatorname{im}}, \ldots, x_4/W_{\operatorname{im}}, y_4/H_{\operatorname{im}}]$.
Suppose the positional features of OCR tokens and objects are $\{p_{1}^{O}, \ldots, p_{o}^{O}\} \subset \mathbb{R}^{8}$, and $\{p_{1}^{D}, \ldots, p_{n}^{D}\} \subset \mathbb{R}^{8}$, respectively, the attended feature is obtained as:
\begin{equation}
\hat{u}_{i}^{P}=\operatorname{Attn}(p_{i}^{O},\{p_{j}^{D}\}_{j=1}^{n},\{u_{j}^{D}\}_{j=1}^{n}).
\end{equation}

Then the final attended object embedding is the sum of them, ${\hat{u}_{i}^{O}=\hat{u}_{i}^{S}+\hat{u}_{i}^{P}}$.

\subsection{Answer Prediction}
The answering component serves to calculate the probability that each answer candidate is an answer to a given question, where each candidate may be either an OCRed text within the image or an additional text. As the questions may not be answered directly using OCRed text, we add some relevant text retrieved by the QA system as additional answer candidates. Following classical QA systems, we use an efficient retrieval system Elasticsearch\footnote{https://github.com/elastic/elasticsearch} to get results related to the question. The QA dataset we use here is composed of (question, answer) pairs in the ST-VQA and TextVQA training sets, although any generic QA dataset could equally be applied. In our experiment, the OCRed text contains only one token or two tokens that are adjacent in the image according to natural reading order of text (i.e., from left to right and top to bottom).

\textbf{Semantic Matching. } In this module, the OCRed text embedding and its attended object embedding are concatenated as the input to a fully connected layer, then we obtain the OCRed answer candidate representation ${u}_{i}^{A}= \operatorname{FC}([{{u}}_{i}^{O};\hat{{u}}_{i}^{O}])$.  We compute the probability that the $i$-th OCRed text is the answer to the question,
\begin{equation}
P_{i}^{A} = \operatorname{softmax} ((u^{Q})^{T} W_{A} u_{i}^{A}),
\end{equation}
where $u^{Q}$ is the question vector and $W_A$ is a parameterized matrix.

\textbf{Semantic Reasoning. } At times, the answer to the question is not composed of the OCR tokens within the image, but needs to be inferred based on the OCRed text. Therefore, in this module, OCRed texts are used to predict possible answers from the additional texts retrieved by the QA system. Specifically, we fuse the OCRed text probabilities into the computation of additional text probability via a GRU, $t^{Q}=\operatorname{GRU}(u^{Q}, \sum_{i} P_{i}^{A} u_{i}^{A})$. And the probability that the answer should be the $j$-th additional text is:
\begin{equation}
P_{j}^{AA} = \operatorname{softmax} ((t^{Q})^{T} W_{AA} u_{j}^{AA}),
\end{equation}
where $W_{AA}$ is a parameterized matrix and $u_{j}^{AA}$ is the $j$-th additional text vector, which is obtained in the same way as OCRed text.

Moreover, for text-based VQA dataset, the answer could also be affirmation ``yes", negation ``no" or no answer ``unanswerable". We separately generate three probabilities $P_Y, P_N, P_U$ corresponding to these three scenarios respectively, following the approach in SDNet \cite{zhu2018sdnet}. For instance, to generate the probability that the answer is ``yes", $P_Y$ , we use:
\begin{equation}
\begin{array}{c}
P_{i}^{Y} = \operatorname{softmax}(({u}^{Q})^{T} W_{Y} {u}_{i}^{A})\\
P_{Y}=(\sum_{i} P_{i}^{Y} {u}_{i}^{A})^{T} {w}_{Y}
\end{array}
\end{equation}
where $W_{Y}$ and ${w}_{Y}$ are the parameterized matrix and vector, respectively. Finally, we select the text with the highest probability from above answer candidates as the final answer to the given question. The binary cross entropy loss is employed here as the objective function to train the model.
\section{Experiments}
\subsection{Datasets and Evaluation Metrics}
We evaluate our RUArt on ST-VQA (Scene Text Visual Question Answering) \cite{biten2019scene} and TextVQA \cite{singh2019towards} datasets, where questions can be answered based on the text embedded in the image. The ST-VQA challenge was structured as 3 tasks of increasing difficulty. The local dictionaries task (Task 1) provides for each image a different dictionary of 100 words that includes the correct answer among a number of distractors. The open dictionary task (Task 3) is the most generic and challenging one among all the the three tasks, since no dictionary is provided \cite{biten2019icdar}. In ST-VQA dataset, the majority of our experiments are conducted on Task 3. The ST-VQA dataset comprises images from different standard datasets that contain scene text, such as COCO-Text \cite{veit2016coco}, VizWiz \cite{gurari2018vizwiz}, ICDAR 2013 \cite{Karatzas2013ICDAR}, ICDAR 2015 \cite{Karatzas2015ICDAR} and IIIT Scene Text Retrieval \cite{Mishra2013Image} dataset, as well as images from generic datasets such as ImageNet \cite{Deng2009ImageNet} and Visual Genome \cite{Krishna2017Visual}, where each selected image contains at least two text instances. ST-VQA contains about 23k images with up to three question-answers pairs per image, and is split into the training set (about 19k images and 26k QA pairs) and the test set (about 3k images and 4k QA pairs per task). The training and validation sets of TextVQA are collected from the training set of the Open Images v3 dataset \cite{openimages}, while the test set is collected from the Open Images' test set. TextVQA contains about 28k images with up to two question-answers pairs per image, and is split into the training set (about 22k images and 34k QA pairs), the validation set (about 3k images and 5k QA pairs) and the test set (about 3k images and 5k QA pairs).

The evaluation metric in TextVQA is the same as the VQA accuracy metric \cite{GoyalKSBP17}, i.e.,
\begin{equation}
\operatorname{Acc}(a n s)=\min \{\frac{\# \text { humans that said } a n s}{3}, 1\}
\end{equation}
The evaluation metric in ST-VQA is the ANLS (average normalized Levenshtein similarity) \cite{biten2019icdar},
\begin{equation}
\text{ANLS}=\frac{1}{N} \sum_{i=0}^{N}\left(\max _{j} s(a_{i j}, o_{q_{i}})\right)
\vspace{-0.2cm}
\end{equation}
$$\begin{aligned}
s\left(a_{i j}, o_{q_{i}}\right)=\left\{\begin{array}{ll}{1-\mathrm{NL}\left(a_{i j}, o_{q_{i}}\right)} & {\text{if } \mathrm{NL}\left(a_{i j}, o_{q_{i}}\right)<\tau} \\ {0} & {\text{if } \mathrm{NL}\left(a_{i j}, o_{q_{i}}\right) \geqslant \tau}\end{array}\right.
\end{aligned}$$
where $N$ is the total number of questions, $M$ is the total number of GT answers per question, $a_{i j}$ ($0 \leq i \leq N,~0 \leq j \leq M$) is the ground truth answer, $o_{q_{i}}$ is the network's answer for the $i^{th}$ question $q_{i}$, $ \mathrm{NL}(a_{i j}, o_{q_{i}})$ is the normalized Levenshtein distance between the strings $a_{i j}$ and $o_{q_{i}}$, and $\tau = 0.5$.
In our experiments, only the results of TextVQA are evaluated using the VQA accuracy metric, and the other results are evaluated using the ANLS metric.

\subsection{Implementation Details}
The implementation of RUArt is based on PyTorch. We utilize the Adamax \cite{Kingma2014Adam} optimizer with a batch size of 16, and the initial learning rate is set to 2e-3 and the weight decay is 0. The number of epochs is set to 30. All the experiments are conducted on one NVIDIA GTX 1080Ti GPU.

\subsection{Ablation Studies}
As text and objects within the images are the basis of RUArt, we first conduct extensive experiments on ST-VQA data sets to explore the performance of various different OCR and object detection models used in RUArt\footnote{For fair comparison and easy repetition, all OCR models and object detection models used in our experiments are pre-trained open-source models.} (shown in Table \ref{tab1}).

\renewcommand\arraystretch{1.2}
\begin{table}
\centering
\newcommand{\tabincell}[2]{\begin{tabular}{@{}#1@{}}#2\end{tabular}}
\caption{Comparison of different OCR and object detection models on ST-VQA Task 3 test set with the metric ANLS.}
\vskip -1mm
\begin{tabular}{p{1.20cm}p{1.35cm}p{1.1cm}p{1.1cm}p{1.1cm}}
\toprule
\multicolumn{1}{c}{Models}&\tabincell{c}{Mask\\TextSpotter} & \tabincell{c}{~PMTD-\\~MORAN}  &\tabincell{c}{~CRAFT-\\~ASTER} &\tabincell{c}{~PMTD-\\~ASTER}\\
  \midrule
\multicolumn{1}{c}{Yolov3} & \multicolumn{1}{c}{0.2807} & \multicolumn{1}{c}{~0.2703} & \multicolumn{1}{c}{0.2840}  & \multicolumn{1}{c}{0.2887}\\ \hline
  \multicolumn{1}{c}{Bottom-Up} & \multicolumn{1}{c}{0.2868} & \multicolumn{1}{c}{~0.2800} & \multicolumn{1}{c}{0.2882}& \multicolumn{1}{c}{\textbf{0.2931}} \\
   \bottomrule
  \end{tabular}
  \label{tab1}
\vspace{-0.4cm}
\end{table}

\textbf{OCR Models: } We firstly adopt an end-to-end scene text spotting method, Mask TextSpotter \cite{lyu2018mask}.  In addition, PMTD \cite{liu2019pyramid} and CRAFT \cite{baek2019character} models are used for text detection, while MORAN \cite{luo2019a} and ASTER \cite{shi2019aster} models are used for text recognition.  In Table \ref{tab1}, we compare the performance of these models on the test set of ST-VQA (task3). The text detection and recognition models are presented in the first row. For instance, PMTD-MORAN represents a two-stage text recognition method, where the model PMTD \cite{liu2019pyramid} is used in the text detection stage and the model MORAN \cite{luo2019a} is used in the text recognition stage. As can been seen from Table \ref{tab1}, the OCRed text obtained by different OCR models has a great impact on the performance of the subsequent text-based VQA model, and PMTD-ASTER obtains the best performance (0.2931 and 0.2887) with different object detection models.

\textbf{Object Detection Models: } We use the pre-trained open source yolov3 model \cite{darknet13} and the bottom-up attention model \cite{anderson2018bottom} to obtain objects in images. The bottom-up attention model, based on Faster R-CNN \cite{ren2015faster} with ResNet-101 \cite{DBLP:conf/cvpr/HeZRS16}, are trained on ImageNet \cite{Russakovsky2015ImageNet} and Visual Genome \cite{Krishna2017Visual} for predicting attribute classes in addition to object classes, such as ``red bus". In Table \ref{tab1}, we compare the performance of these two object detection models on the test set of ST-VQA (task3). The bottom-up attention model performs better than the yolov3 model with any OCR model.

Next, we perform a number of ablation studies (shown in Table \ref{tab2}) to analyze the reasons for the improvements of our RUArt.  We use RUArt-base with PMTD-ASTER (the OCR model) and the bottom-up attention method (the object detection model) as the initial model, and we use the OCRed text of ``1 or 2 tokens" as answers. Here, the training data is the ST-VQA training set.

\renewcommand\arraystretch{1.3}
\begin{table}
\centering
\caption{Ablation studies of RUArt on ST-VQA Task 3 test set with the metric ANLS, where ``-" means removing the module from RUArt.}
\vskip -2mm
\begin{tabular}{ p{6cm}c}
Model & ANLS \\ \toprule[1pt]
RUArt-base                      & 0.2931     \\ \hline
Input Component&                                \\
\quad+TextVQA Training Data&         0.3108           \\
\quad Only 1-token answers &0.2750 \\ \hline
OCR and Object Understanding            &                                \\
\quad-Word-level Attention      &     0.2882      \\
\quad-Multilevel Attention      &    0.2867      \\
\quad-Self-Attention            &     0.2879        \\
\quad-Above Three Attentions     &     0.2838  \\ \hline
Relational Reasoning     &                     \\
\quad-Semantic Attention        &     0.2896              \\
\quad-Positional Attention      &     0.2877            \\
\quad Replacing Relational Reasoning with Object& \multirow{2}{*}{0.2865}  \\
Weighted Sum  & \\ \hline
Answer Prediction     &                                \\
\quad-Attended Object Embedding&      0.2854               \\
\quad-OCRed Text Embedding      & 0.2890\\
\quad+Semantic Reasoning & \textbf{0.3133}\\
\toprule[1pt]
\end{tabular}
\label{tab2}
\vspace{-0.4cm}
\end{table}

\textbf{Input Component:} First, we augment ST-VQA data set with TextVQA dataset to train the RUArt model, and the performance is improved from 0.2931 to 0.3108.  Next, all other ablation experiments (except the previous data augmentation experiment) are trained only on the ST-VQA data set. In our initial experiment, we select the OCRed text containing 1 or 2 tokens extracted from images as answers. If we only use 1-token OCRed text as answers, the performance is reduced to 0.2750. Of course, we can also add the OCRed text including 3 or more tokens to the answer candidates. But for the balance of performance and complexity, we only consider answer candidates of 1 or 2 OCR tokens.

\textbf{OCR and Object Understanding:} The OCR and object understanding module employs several attention mechanism to integrate the information of OCR tokens, objects and the question. The word-level attention is used to learn the initial relationship between each OCR token/object and the question, the multilevel attention is used to learn the relationships between different layers of the OCR token/object and the question, and the self-attention is used to establish direct correlations between all pairs of OCR tokens/objects. As shown in Table \ref{tab2}, when any attention part is removed, the performance is somewhat reduced. When three attention parts are removed simultaneously, the performance drops to 0.2838.

\textbf{Relational Reasoning:} In the relational reasoning module, we conduct positional attention and semantic attention from objects to OCR tokens, respectively. As illustrated in Table \ref{tab2}, when the semantic attention is removed, the performance drops to 0.2896, while when the positional attention is removed, the performance drops to 0.2877. These results show that the positional relations are a little more important than the semantic relations between OCR tokens and objects in the image for text-based VQA. When we replace OCR-object relational reasoning with the object weighted sum, where the weight of $i$-th object is $\alpha_{i} \propto \exp (w^{T} u_{i}^{D})$ and $w$ is a parameterized vector, the performance drops from 0.2931 to 0.2865. This indicates that in the use of the object information, finding the relationships between OCR tokens and objects is important.

\textbf{Answer Prediction:} As mentioned above, our RUArt-base only selects answers from the OCRed text, so we first perform ablation experiments that remove OCRed text embedding and attended object embedding respectively in semantic matching. As can be seen from Table \ref{tab2}, removing any of these embeddings may reduce the performance, and the performance drops more when ignoring the attended object embedding. This is because the attended object embedding learns the relationships between OCR tokens and objects, while the majority of questions involve both text and objects. Next, we retrieve the answers related to the question by the QA system and add them as additional text to the answer candidates. Finally, we adopt semantic reasoning, which infers answers from additional text based on OCR tokens, to answer questions that cannot be directly answered with OCRed text. And we achieve a score of 0.3133, which is around 2\% (absolute) higher than that of RUArt-base.

\subsection{Results and Analysis}
In this section, we compare our RUArt\footnote{The inference model of our RUArt is available at https://github.com/xiaojino/RUArt.} model with the state-of-the-art methods on ST-VQA and TextVQA. In the experiment of ST-VQA, we use the PMTD \cite{liu2019pyramid} text detection model, the ASTER \cite{shi2019aster} text recognition model and the Bottom-Up object detection \cite{anderson2018bottom}  model. The training data of ST-VQA is used to train the model. And answer candidates are from the OCRed text of  ``1 or 2 OCR tokens" or additional text (including `yes', `no', `unanswerable' and top-10 retrieved results).

\begin{table}
\centering
\caption{Comparison with participants of ST-VQA on test set with the metric ANLS.}
\vskip -1mm
\begin{tabular}{cccc}
\toprule
Methods & Task 1 & Task 3\\
\midrule
USTB-TQA \cite{biten2019icdar} & 0.455  & 0.170\\
Clova AI OCR \cite{biten2019icdar} & -  & 0.215\\
QAQ \cite{biten2019icdar} & -  & 0.256 \\
MM-GNN \cite{Gao_2020_CVPR} & -  & 0.207 \\
VTA \cite{biten2019icdar} & 0.506  & 0.282 \\
M4C \cite{Hu_2020_CVPR} & -  & 0.462 \\ \hline
RUArt & 0.482  & 0.313\\
RUArt-M4C & -  & \textbf{0.481}\\
\bottomrule
\end{tabular}
\label{tab3}
\vspace{-0.4cm}
\end{table}

\begin{figure*}
  \centering
  \includegraphics[width=0.98\textwidth]{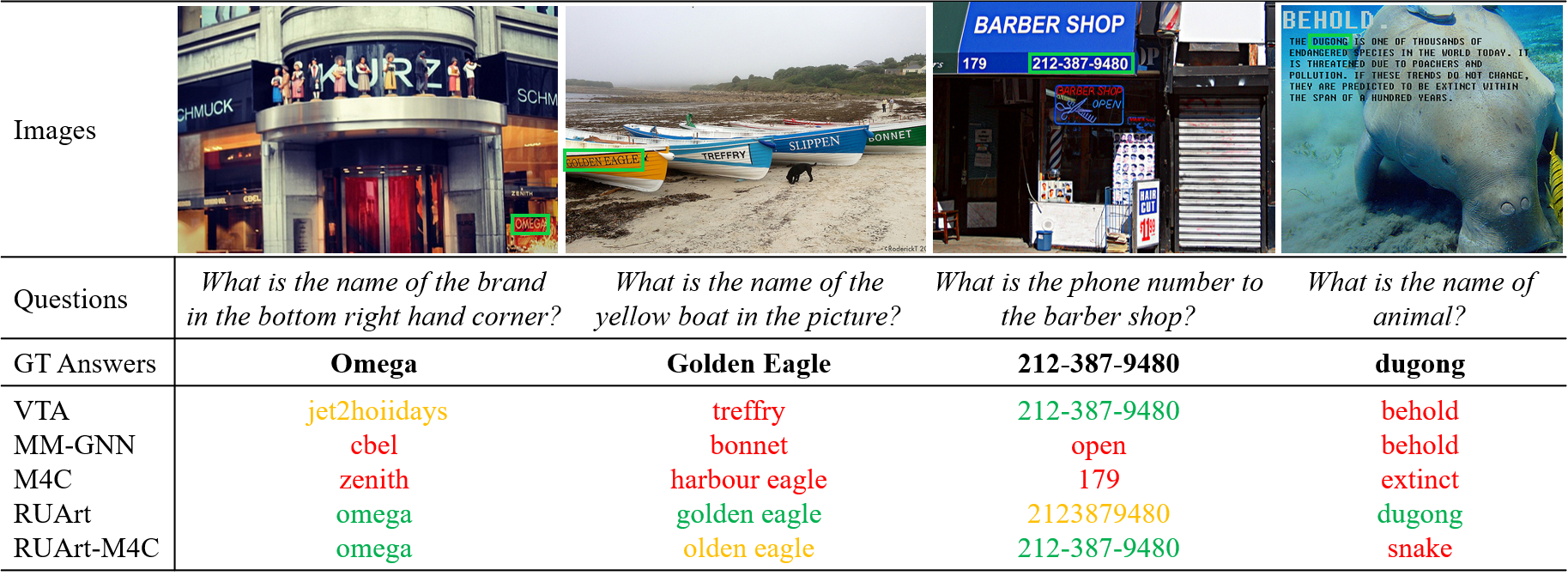}
  \vspace{-0.2cm}
  \caption{Qualitative comparison of RUArt and other participants on the ST-VQA Task 3 test set. Green text indicates that the answer is correct, red text indicates that the answer is wrong, and orange text indicates that the answer is partially correct. Our model presents great performance in relational reasoning, text semantic understanding and reasoning.}
  \label{example}
  \vspace{-0.4cm}
\end{figure*}

From the results in Table \ref{tab3}, we can see that RUArt (with ANLS 0.313) obtains an obvious improvement over the majority of state-of-the-art methods on the ST-VQA Task 3 test set.  Notably, the VTA method is the winner of the ST-VQA challenge 2019. And we also conduct experiment on Task 1 to further verify the performance of our method. As the answer candidates dictionary for each image is provided in Task 1, we directly replace the OCR tokens with the dictionary (its positions are filled with 0) for experiment. It achieves a score of 0.482, which is lower than that of the VTA.  One of the reason is that the provided dictionary is generated by multiple methods \cite{biten2019icdar}, and it cannot form a reasonable text context, which affects the learning of the contextual information of OCR tokens. On the other hand, the positions of the text in the image are not provided, so the relationship between the text and objects based on the positions cannot be explored.

And our method is lower than the M4C, it is affected by many factors. For example, M4C replaces many low-quality images from COCO-Text (around 1/3 of all images) with high-quality images as inputs to object detection and OCR systems \cite{Hu_2020_CVPR}, while our method still use the images provided in ST-VQA. And M4C predicts the answer for less than 13 OCR tokens, while our method only predicts the answer of 1 or 2 OCR tokens. It is worth noting that our method focuses on text-centered semantic understanding  and reasoning, and it is not mutually exclusive but complementary to other multi-modal methods. Therefore, in order to prove the generality and effectiveness of our method, we simply add the key modules in RUArt (i.e., OCR understanding, relational reasoning, semantic matching and reasoning) to the M4C model for experiments. Specifically, we first add the OCR embeddings obtained through text understanding and relational reasoning as the OCR text feature to M4C method. Then, the semantic matching between the question and the OCR tokens is added to enhance the guiding role of the question semantics in the answer prediction stage. Finally, the semantic reasoning based on OCR tokens is added to select the answer from the fixed dictionary. Consequently, this improved method (marked as RUArt-M4C) achieves a score of 0.481 (shown in Table \ref{tab3}), which is around 2\% (absolute) higher than that of M4C.

We also present some qualitative samples of ST-VQA Task 3 test set in Fig. \ref{example}, which indicates that our method performs better than others in relational reasoning, text semantic understanding and reasoning. The first two examples require mining the relationship between the text and the object. It can be seen that RUArt has excellent performance in relational reasoning. In addition, when the RUArt's key modules are added to the M4C method (marked as RUArt-M4C), the prediction errors of the M4C on such questions can also be corrected. The last two samples require text understanding, semantic matching and reasoning. Other methods select the prominent or frequently appearing words in the dataset as the answers without understanding the question. On the contrary, RUArt achieves better performance when answering such questions, and it can even be learned that ``dugong" is an animal in the last case. As for the RUArt-M4C method, although the ``dugong" is not found correctly, the selected answer ``snake" also belongs to the animal, which also verifies that our modules are really helpful for the understanding of the question.

\begin{table}
\centering
\caption{Comparison of RUArt and RUArt* on different subsets of ST-VQA Task 3 test set, where the number in bracket is the accuracy (\%) of OCR.}
\vskip -1mm
\begin{tabular}{p{1cm}p{1cm}p{1cm}p{1cm}p{0.8cm}}
\toprule
Data &\multicolumn{1}{c}{COCO-Text}&\multicolumn{1}{c}{ICDAR}&Other&Total\\
Ratio &\multicolumn{1}{c}{34\%}&\multicolumn{1}{c}{5\%}&61\%&100\%\\
\midrule
RUArt&\multicolumn{1}{c}{0.169~(5.9)}& \multicolumn{1}{c}{0.465~(90.5)} &0.381&0.313 \\
RUArt*&\multicolumn{1}{c}{0.538~(100)}& \multicolumn{1}{c}{0.487~(100)} &0.379&0.438 \\
\bottomrule
\end{tabular}
\label{tab4}
\vspace{-0.4cm}
\end{table}
\begin{figure}
  \centering
  \includegraphics[width=0.46\textwidth]{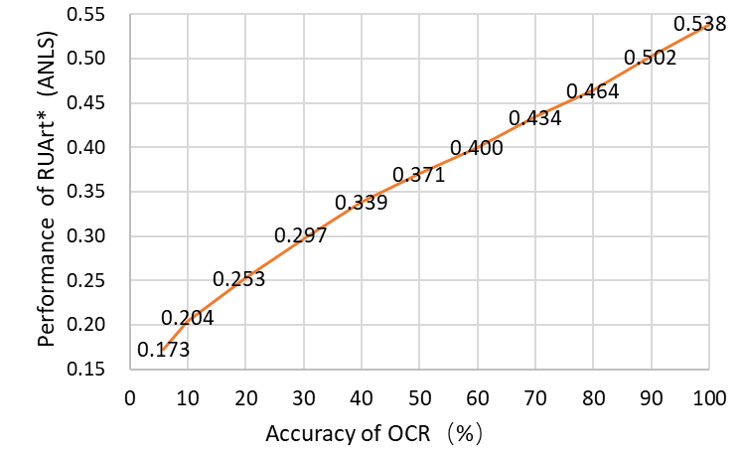}
  \vspace{-0.2cm}
  \caption{Performances of RUArt* with different accuracy of OCR on the COCO-Text subset of ST-VQA Task 3 test set.}
  \label{performance}
  \vspace{-0.6cm}
\end{figure}

In order to quantitatively evaluate the impact of OCRed text on a text-based VQA model, we add 38\% and 39\% ground truth of OCR tokens (from COCO-Text and ICDAR data) to the training set and test set respectively, to train and conduct inference for the model,  marked as RUArt*. It achieves a surprised score of 0.438, which is 40\% higher than that of the RUArt. To analyze the specific reasons for the large improvement, we compare RUArt and RUArt* on different sub-sets of the ST-VQA test set, as presented in Table \ref{tab4}. The COCO-Text data accounts for 34\% of the ST-VQA test set, whereas the accuracy of the OCR model (PMTD-ASTER) on the COCO-Text test set is only 5.9\%. This is because many images from COCO-Text in the ST-VQA data are resized to 256 $\times$ 256, which degrades the image quality and distorts their aspect ratios. On the basis of such low-quality OCR results, the RUArt still achieves a performance of 0.169. Without changing the accuracy of the OCR model on the test set, we use RUArt* to conduct inference, and the performance on the COCO-Text subset of test set is improved from 0.169 to 0.173. When we improve the accuracy of the OCR on the COCO-Text subset by random adding some ground truth of OCR tokens, the performance of RUArt* on the COCO-Text subset increases accordingly, as shown in Fig. \ref{performance}. If the accuracy of the OCR reaches 100\%, the RUArt* can achieve the performance of 0.538 on the COCO-Text subset, and the performance on the entire dataset is improved to 0.438. Meanwhile, when the accuracy of OCR on the ICDAR subset is improved from 90.5\% to 100\%, the performance of our method is improved from 0.465 to 0.487. In addition, we can see from Table \ref{tab4} that RUArt also has a good performance on other subsets without any ground truth of OCR.

Finally, we evaluate our model on TextVQA with the accuracy metric. In the experiment of TextVQA, we use the OCR tokens obtained by Rosetta-en, which is provided in its dataset. The training data of TextVQA is used to train the model. And answer candidates are from the OCRed text of  ``1 or 2 OCR tokens" or additional text (including `yes', `no', `unanswerable' and top-10 retrieved results).  Consequently, our RUArt achieves a score of 33.54\%, which also verifies that our proposed method can effectively explore the contextual information of the text and mine the stable relationships between the text and objects.

The performance of M4C method is 5.6\% higher than ours. One of the reasons is that it uses rich OCR representation (including fastText vector, Pyramidal Histogram of Characters vector, appearance feature and location feature), which gives around 4\% (absolute) accuracy improvement compare with using only fastText features \cite{Hu_2020_CVPR}. Another reason is that it uses a iterative decoding method for answer prediction, which leads to around 4\% (absolute) higher accuracy than single-step classifier \cite{Hu_2020_CVPR}. In addition, the classification-based VQA models are prone to overfit the fixed answer pool in TextVQA, which enables impressive performance, but poor generalized to other datasets \cite{Wang_2020_CVPR}. There is no doubt that appearance feature and character feature may introduce new information, and iterative prediction can expand the range of answer candidates, improving the upper bound of method performance. Similarly, fusion of new features and multi-step prediction can further improve the performance of our RUArt, which will be one of our future work.

\begin{table}
\centering
\caption{Comparison with participants of TextVQA on the test set with the metric accuracy (\%).}
\vskip -2mm
\begin{tabular}{cc}
\toprule
Methods & Accuracy(\%)\\
\midrule
Image Only&5.88\\
Pythia &14.01 \\
LoRRA &27.63  \\
Schwail & 30.54 \\
MM-GNN &31.10  \\
DCD\_ZJU (DCD) &31.44  \\
MSFT\_VTI &32.46   \\
M4C & 39.10   \\\hline
RUArt &  33.54\\
\bottomrule
\end{tabular}
\label{tab5}
\vspace{-0.4cm}
\end{table}

\section{Conclusion}
In this paper, we propose a novel text-centered framework (RUArt) for text-based visual question answering. Our approach unifies all the input into the pure text, enhances semantic fusion of different inputs, and makes reasoning more interpretable. In addition, we are the first to explore the relationships between OCR tokens and object properties via semantic and positional attention. Our method achieves comparable performance on current available datasets, i.e., ST-VQA and TextVQA.

In addition to enriching OCR features and making multi-step predictions, the quality of OCR is also very important, as shown in Fig. \ref{performance}. However, individual character missing and recognition errors still exist in current OCR models. Sometimes, OCR tokens can be corrected according to the edit distances from the words in the dictionary. However, if multiple words have the same edit distance from one OCR token, the contextual information must be considered to obtain a more stable OCR result. Therefore, a future work is to add the modification of OCR tokens as a sub-module into our framework, and get better OCR modification results via subsequent-task training.

%\section*{Acknowledgment}

\bibliographystyle{IEEEtran}
\bibliography{IEEEabrv,mybibfile}

\begin{thebibliography}{10}
\providecommand{\url}[1]{#1}
\csname url@samestyle\endcsname
\providecommand{\newblock}{\relax}
\providecommand{\bibinfo}[2]{#2}
\providecommand{\BIBentrySTDinterwordspacing}{\spaceskip=0pt\relax}
\providecommand{\BIBentryALTinterwordstretchfactor}{4}
\providecommand{\BIBentryALTinterwordspacing}{\spaceskip=\fontdimen2\font plus
\BIBentryALTinterwordstretchfactor\fontdimen3\font minus
  \fontdimen4\font\relax}
\providecommand{\BIBforeignlanguage}[2]{{%
\expandafter\ifx\csname l@#1\endcsname\relax
\typeout{** WARNING: IEEEtran.bst: No hyphenation pattern has been}%
\typeout{** loaded for the language `#1'. Using the pattern for}%
\typeout{** the default language instead.}%
\else
\language=\csname l@#1\endcsname
\fi
#2}}
\providecommand{\BIBdecl}{\relax}
\BIBdecl

\bibitem{li2019visual}
H.~Li, P.~Wang, C.~Shen, and A.~van~den Hengel, ``Visual question answering as
  reading comprehension,'' in \emph{{CVPR}}.\hskip 1em plus 0.5em minus
  0.4em\relax Computer Vision Foundation / {IEEE}, 2019, pp. 6319--6328.

\bibitem{yu2019deep}
Z.~Yu, J.~Yu, Y.~Cui, D.~Tao, and Q.~Tian, ``Deep modular co-attention networks
  for visual question answering,'' in \emph{{CVPR}}.\hskip 1em plus 0.5em minus
  0.4em\relax Computer Vision Foundation / {IEEE}, 2019, pp. 6281--6290.

\bibitem{anderson2018bottom}
P.~Anderson, X.~He, C.~Buehler, D.~Teney, M.~Johnson, S.~Gould, and L.~Zhang,
  ``Bottom-up and top-down attention for image captioning and visual question
  answering,'' in \emph{{CVPR}}.\hskip 1em plus 0.5em minus 0.4em\relax {IEEE}
  Computer Society, 2018, pp. 6077--6086.

\bibitem{ben2017mutan}
H.~Ben{-}younes, R.~Cad{\`{e}}ne, M.~Cord, and N.~Thome, ``{MUTAN:} multimodal
  tucker fusion for visual question answering,'' in \emph{{ICCV}}.\hskip 1em
  plus 0.5em minus 0.4em\relax {IEEE} Computer Society, 2017, pp. 2631--2639.

\bibitem{li2019relation}
L.~Li, Z.~Gan, Y.~Cheng, and J.~Liu, ``Relation-aware graph attention network
  for visual question answering,'' in \emph{{ICCV}}.\hskip 1em plus 0.5em minus
  0.4em\relax {IEEE}, 2019, pp. 10\,312--10\,321.

\bibitem{GaoJYLHWL19}
P.~Gao, Z.~Jiang, H.~You, P.~Lu, S.~C.~H. Hoi, X.~Wang, and H.~Li, ``Dynamic
  fusion with intra- and inter-modality attention flow for visual question
  answering,'' in \emph{{CVPR}}.\hskip 1em plus 0.5em minus 0.4em\relax
  Computer Vision Foundation / {IEEE}, 2019, pp. 6639--6648.

\bibitem{biten2019scene}
A.~F. Biten, R.~Tito, A.~Mafla, L.~Gomez, M.~Rusi{\~n}ol, E.~Valveny,
  C.~Jawahar, and D.~Karatzas, ``Scene text visual question answering,'' in
  \emph{{ICCV}}.\hskip 1em plus 0.5em minus 0.4em\relax {IEEE}, 2019, pp.
  4290--4300.

\bibitem{singh2019towards}
A.~Singh, V.~Natarajan, M.~Shah, Y.~Jiang, X.~Chen, D.~Batra, D.~Parikh, and
  M.~Rohrbach, ``Towards {VQA} models that can read,'' in \emph{{CVPR}}.\hskip
  1em plus 0.5em minus 0.4em\relax Computer Vision Foundation / {IEEE}, 2019,
  pp. 8317--8326.

\bibitem{KembhaviSSCFH17}
A.~Kembhavi, M.~J. Seo, D.~Schwenk, J.~Choi, A.~Farhadi, and H.~Hajishirzi,
  ``Are you smarter than a sixth grader? textbook question answering for
  multimodal machine comprehension,'' in \emph{{CVPR}}.\hskip 1em plus 0.5em
  minus 0.4em\relax {IEEE} Computer Society, 2017, pp. 5376--5384.

\bibitem{kafle2018dvqa}
K.~Kafle, B.~L. Price, S.~Cohen, and C.~Kanan, ``{DVQA:} understanding data
  visualizations via question answering,'' in \emph{{CVPR}}.\hskip 1em plus
  0.5em minus 0.4em\relax {IEEE} Computer Society, 2018, pp. 5648--5656.

\bibitem{antol2015vqa}
S.~Antol, A.~Agrawal, J.~Lu, M.~Mitchell, D.~Batra, C.~L. Zitnick, and
  D.~Parikh, ``{VQA:} visual question answering,'' in \emph{{ICCV}}.\hskip 1em
  plus 0.5em minus 0.4em\relax {IEEE} Computer Society, 2015, pp. 2425--2433.

\bibitem{biten2019icdar}
A.~F. Biten, R.~Tito, A.~Mafla, L.~Gomez, M.~Rusi{\~n}ol, M.~Mathew,
  C.~Jawahar, E.~Valveny, and D.~Karatzas, ``{ICDAR} 2019 competition on scene
  text visual question answering,'' in \emph{{ICDAR}}.\hskip 1em plus 0.5em
  minus 0.4em\relax {IEEE}, 2019, pp. 1563--1570.

\bibitem{devlin2018bert}
J.~Devlin, M.~Chang, K.~Lee, and K.~Toutanova, ``{BERT:} pre-training of deep
  bidirectional transformers for language understanding,'' in
  \emph{{NAACL-HLT}}.\hskip 1em plus 0.5em minus 0.4em\relax Association for
  Computational Linguistics, 2019, pp. 4171--4186.

\bibitem{Gao_2020_CVPR}
D.~Gao, K.~Li, R.~Wang, S.~Shan, and X.~Chen, ``Multi-modal graph neural
  network for joint reasoning on vision and scene text,'' in \emph{IEEE/CVF
  Conference on Computer Vision and Pattern Recognition (CVPR)}, June 2020.

\bibitem{Hu_2020_CVPR}
R.~Hu, A.~Singh, T.~Darrell, and M.~Rohrbach, ``Iterative answer prediction
  with pointer-augmented multimodal transformers for textvqa,'' in
  \emph{IEEE/CVF Conference on Computer Vision and Pattern Recognition (CVPR)},
  June 2020.

\bibitem{zhu2018sdnet}
C.~Zhu, M.~Zeng, and X.~Huang, ``{SDNet:} contextualized attention-based deep
  network for conversational question answering,'' \emph{arXiv preprint
  arXiv:1812.03593}, 2018.

\bibitem{yang2016stacked}
Z.~Yang, X.~He, J.~Gao, L.~Deng, and A.~J. Smola, ``Stacked attention networks
  for image question answering,'' in \emph{{CVPR}}.\hskip 1em plus 0.5em minus
  0.4em\relax {IEEE} Computer Society, 2016, pp. 21--29.

\bibitem{fukui2016multimodal}
A.~Fukui, D.~H. Park, D.~Yang, A.~Rohrbach, T.~Darrell, and M.~Rohrbach,
  ``Multimodal compact bilinear pooling for visual question answering and
  visual grounding,'' in \emph{{EMNLP}}.\hskip 1em plus 0.5em minus 0.4em\relax
  The Association for Computational Linguistics, 2016, pp. 457--468.

\bibitem{kim2016hadamard}
J.~Kim, K.~W. On, W.~Lim, J.~Kim, J.~Ha, and B.~Zhang, ``Hadamard product for
  low-rank bilinear pooling,'' in \emph{{ICLR}}, 2017.

\bibitem{yu2017multi}
Z.~Yu, J.~Yu, J.~Fan, and D.~Tao, ``Multi-modal factorized bilinear pooling
  with co-attention learning for visual question answering,'' in
  \emph{{ICCV}}.\hskip 1em plus 0.5em minus 0.4em\relax {IEEE} Computer
  Society, 2017, pp. 1839--1848.

\bibitem{Liang0CLH18}
J.~Liang, L.~Jiang, L.~Cao, L.~Li, and A.~G. Hauptmann, ``Focal visual-text
  attention for visual question answering,'' in \emph{CVPR}.\hskip 1em plus
  0.5em minus 0.4em\relax {IEEE} Computer Society, 2018, pp. 6135--6143.

\bibitem{yu2018beyond}
Z.~Yu, J.~Yu, C.~Xiang, J.~Fan, and D.~Tao, ``Beyond bilinear: Generalized
  multimodal factorized high-order pooling for visual question answering,''
  \emph{{IEEE} Trans. Neural Netw. Learning Syst.}, vol.~29, no.~12, pp.
  5947--5959, 2018.

\bibitem{hudson2018compositional}
D.~A. Hudson and C.~D. Manning, ``Compositional attention networks for machine
  reasoning,'' in \emph{{ICLR}}.\hskip 1em plus 0.5em minus 0.4em\relax
  OpenReview.net, 2018.

\bibitem{chen2017reading}
D.~Chen, A.~Fisch, J.~Weston, and A.~Bordes, ``Reading wikipedia to answer
  open-domain questions,'' in \emph{{ACL}}.\hskip 1em plus 0.5em minus
  0.4em\relax Association for Computational Linguistics, 2017, pp. 1870--1879.

\bibitem{jin2019health}
Z.~Jin, B.~Zhang, F.~Fang, L.~Zhang, and X.~Yin, ``Health assistant: answering
  your questions anytime from biomedical literature,'' \emph{Bioinformatics},
  vol.~35, no.~20, pp. 4129--4139, 2019.

\bibitem{Jin2020Ranking}
Z.~Jin, B.~Zhang, F.~Zhou, J.~Qin, and X.~Yin, ``Ranking via partial ordering
  for answer selection,'' \emph{Information Sciences}, vol. 538C, pp. 358--371,
  2020.

\bibitem{chen2012understanding}
N.~Chen, Q.~Zhou, and V.~K. Prasanna, ``Understanding web images by object
  relation network,'' in \emph{{WWW}}.\hskip 1em plus 0.5em minus 0.4em\relax
  {ACM}, 2012, pp. 291--300.

\bibitem{hu2018relation}
H.~Hu, J.~Gu, Z.~Zhang, J.~Dai, and Y.~Wei, ``Relation networks for object
  detection,'' in \emph{{CVPR}}.\hskip 1em plus 0.5em minus 0.4em\relax {IEEE}
  Computer Society, 2018, pp. 3588--3597.

\bibitem{zhou2018object}
L.~Zhou, J.~Zhao, J.~Li, L.~Yuan, and J.~Feng, ``Object relation detection
  based on one-shot learning,'' \emph{arXiv preprint arXiv:1807.05857}, 2018.

\bibitem{song2019image}
X.~Song, S.~Jiang, B.~Wang, C.~Chen, and G.~Chen, ``Image representations with
  spatial object-to-object relations for rgb-d scene recognition,''
  \emph{{IEEE} Trans. Image Processing}, vol.~29, pp. 525--537, 2019.

\bibitem{Peng0WWH19}
L.~Peng, Y.~Yang, Z.~Wang, X.~Wu, and Z.~Huang, ``Cra-net: Composed relation
  attention network for visual question answering,'' in \emph{{ACM}
  {MM}}.\hskip 1em plus 0.5em minus 0.4em\relax {ACM}, 2019, pp. 1202--1210.

\bibitem{HanSLYS18}
C.~Han, F.~Shen, L.~Liu, Y.~Yang, and H.~T. Shen, ``Visual spatial attention
  network for relationship detection,'' in \emph{{ACM} {MM}}.\hskip 1em plus
  0.5em minus 0.4em\relax {ACM}, 2018, pp. 510--518.

\bibitem{JinZG00Z19}
W.~Jin, Z.~Zhao, M.~Gu, J.~Yu, J.~Xiao, and Y.~Zhuang, ``Multi-interaction
  network with object relation for video question answering,'' in \emph{{ACM}
  {MM}}.\hskip 1em plus 0.5em minus 0.4em\relax {ACM}, 2019, pp. 1193--1201.

\bibitem{Singh0SC19}
A.~K. Singh, A.~Mishra, S.~Shekhar, and A.~Chakraborty, ``From strings to
  things: Knowledge-enabled {VQA} model that can read and reason,'' in
  \emph{{ICCV}}.\hskip 1em plus 0.5em minus 0.4em\relax {IEEE}, 2019, pp.
  4601--4611.

\bibitem{Graves1997Long}
S.~Hochreiter and J.~Schmidhuber, ``Long short-term memory,'' \emph{Neural
  Computation}, vol.~9, no.~8, pp. 1735--1780, 1997.

\bibitem{vaswani2017attention}
A.~Vaswani, N.~Shazeer, N.~Parmar, J.~Uszkoreit, L.~Jones, A.~N. Gomez,
  L.~Kaiser, and I.~Polosukhin, ``Attention is all you need,'' in
  \emph{{NIPS}}, 2017, pp. 5998--6008.

\bibitem{Pennington2014Glove}
J.~Pennington, R.~Socher, and C.~D. Manning, ``Glove: Global vectors for word
  representation,'' in \emph{{EMNLP}}.\hskip 1em plus 0.5em minus 0.4em\relax
  {ACL}, 2014, pp. 1532--1543.

\bibitem{HuangZSC18}
H.~Huang, C.~Zhu, Y.~Shen, and W.~Chen, ``Fusionnet: Fusing via fully-aware
  attention with application to machine comprehension,'' in
  \emph{{ICLR}}.\hskip 1em plus 0.5em minus 0.4em\relax OpenReview.net.

\bibitem{GraveMJB17}
A.~Joulin, E.~Grave, P.~Bojanowski, and T.~Mikolov, ``Bag of tricks for
  efficient text classification,'' in \emph{{EACL}}, M.~Lapata, P.~Blunsom, and
  A.~Koller, Eds.\hskip 1em plus 0.5em minus 0.4em\relax Association for
  Computational Linguistics, 2017, pp. 427--431.

\bibitem{veit2016coco}
A.~Veit, T.~Matera, L.~Neumann, J.~Matas, and S.~Belongie, ``Coco-text: Dataset
  and benchmark for text detection and recognition in natural images,''
  \emph{arXiv preprint arXiv:1601.07140}, 2016.

\bibitem{gurari2018vizwiz}
D.~Gurari, Q.~Li, A.~J. Stangl, A.~Guo, C.~Lin, K.~Grauman, J.~Luo, and J.~P.
  Bigham, ``Vizwiz grand challenge: Answering visual questions from blind
  people,'' in \emph{{CVPR}}.\hskip 1em plus 0.5em minus 0.4em\relax {IEEE}
  Computer Society, 2018, pp. 3608--3617.

\bibitem{Karatzas2013ICDAR}
D.~Karatzas, F.~Shafait, S.~Uchida, M.~Iwamura, L.~G. i~Bigorda, S.~R. Mestre,
  J.~Mas, D.~F. Mota, J.~Almaz{\'{a}}n, and L.~de~las Heras, ``{ICDAR} 2013
  robust reading competition,'' in \emph{{ICDAR}}.\hskip 1em plus 0.5em minus
  0.4em\relax {IEEE} Computer Society, 2013, pp. 1484--1493.

\bibitem{Karatzas2015ICDAR}
D.~Karatzas, L.~Gomez{-}Bigorda, A.~Nicolaou, S.~K. Ghosh, A.~D. Bagdanov,
  M.~Iwamura, J.~Matas, L.~Neumann, V.~R. Chandrasekhar, S.~Lu, F.~Shafait,
  S.~Uchida, and E.~Valveny, ``{ICDAR} 2015 competition on robust reading,'' in
  \emph{{ICDAR}}.\hskip 1em plus 0.5em minus 0.4em\relax {IEEE} Computer
  Society, 2015, pp. 1156--1160.

\bibitem{Mishra2013Image}
A.~Mishra, K.~Alahari, and C.~V. Jawahar, ``Image retrieval using textual
  cues,'' in \emph{{ICCV}}.\hskip 1em plus 0.5em minus 0.4em\relax {IEEE}
  Computer Society, 2013, pp. 3040--3047.

\bibitem{Deng2009ImageNet}
J.~Deng, W.~Dong, R.~Socher, L.~Li, K.~Li, and F.~Li, ``Imagenet: {A}
  large-scale hierarchical image database,'' in \emph{{CVPR}}.\hskip 1em plus
  0.5em minus 0.4em\relax {IEEE} Computer Society, 2009, pp. 248--255.

\bibitem{Krishna2017Visual}
R.~Krishna, Y.~Zhu, O.~Groth, J.~Johnson, K.~Hata, J.~Kravitz, S.~Chen,
  Y.~Kalantidis, L.~Li, D.~A. Shamma, M.~S. Bernstein, and L.~Fei{-}Fei,
  ``Visual genome: Connecting language and vision using crowdsourced dense
  image annotations,'' \emph{International Journal of Computer Vision}, vol.
  123, no.~1, pp. 32--73, 2017.

\bibitem{openimages}
I.~Krasin, T.~Duerig, N.~Alldrin, V.~Ferrari, S.~Abu-El-Haija, A.~Kuznetsova,
  H.~Rom, J.~Uijlings, S.~Popov, A.~Veit, S.~Belongie, V.~Gomes, A.~Gupta,
  C.~Sun, G.~Chechik, D.~Cai, Z.~Feng, D.~Narayanan, and K.~Murphy,
  ``Openimages: A public dataset for large-scale multi-label and multi-class
  image classification.'' \emph{Dataset available from
  https://github.com/openimages}, 2017.

\bibitem{GoyalKSBP17}
Y.~Goyal, T.~Khot, D.~Summers{-}Stay, D.~Batra, and D.~Parikh, ``Making the {V}
  in {VQA} matter: Elevating the role of image understanding in visual question
  answering,'' in \emph{{CVPR}}.\hskip 1em plus 0.5em minus 0.4em\relax {IEEE}
  Computer Society, 2017, pp. 6325--6334.

\bibitem{Kingma2014Adam}
D.~P. Kingma and J.~Ba, ``Adam: {A} method for stochastic optimization,'' in
  \emph{{ICLR}}, 2015.

\bibitem{lyu2018mask}
P.~Lyu, M.~Liao, C.~Yao, W.~Wu, and X.~Bai, ``Mask textspotter: An end-to-end
  trainable neural network for spotting text with arbitrary shapes,'' in
  \emph{{ECCV}}, ser. Lecture Notes in Computer Science, vol. 11218.\hskip 1em
  plus 0.5em minus 0.4em\relax Springer, 2018, pp. 71--88.

\bibitem{liu2019pyramid}
J.~Liu, X.~Liu, J.~Sheng, D.~Liang, X.~Li, and Q.~Liu, ``Pyramid mask text
  detector,'' \emph{arXiv preprint arXiv:1903.11800}, 2019.

\bibitem{baek2019character}
Y.~Baek, B.~Lee, D.~Han, S.~Yun, and H.~Lee, ``Character region awareness for
  text detection,'' in \emph{{CVPR}}.\hskip 1em plus 0.5em minus 0.4em\relax
  Computer Vision Foundation / {IEEE}, 2019, pp. 9365--9374.

\bibitem{luo2019a}
C.~Luo, L.~Jin, and Z.~Sun, ``{MORAN:} {A} multi-object rectified attention
  network for scene text recognition,'' \emph{Pattern Recognition}, vol.~90,
  pp. 109--118, 2019.

\bibitem{shi2019aster}
B.~Shi, M.~Yang, X.~Wang, P.~Lyu, C.~Yao, and X.~Bai, ``{ASTER:} an attentional
  scene text recognizer with flexible rectification,'' \emph{{IEEE} {TPAMI}},
  vol.~41, no.~9, pp. 2035--2048, 2019.

\bibitem{darknet13}
J.~Redmon, ``Darknet: Open source neural networks in c,''
  \url{http://pjreddie.com/darknet/}, 2013--2016.

\bibitem{ren2015faster}
S.~Ren, K.~He, R.~B. Girshick, and J.~Sun, ``Faster {R-CNN:} towards real-time
  object detection with region proposal networks.'' \emph{IEEE TPAMI}, vol.~39,
  no.~6, pp. 1137--1149, 2015.

\bibitem{DBLP:conf/cvpr/HeZRS16}
K.~He, X.~Zhang, S.~Ren, and J.~Sun, ``Deep residual learning for image
  recognition,'' in \emph{{CVPR}}.\hskip 1em plus 0.5em minus 0.4em\relax
  {IEEE} Computer Society, 2016, pp. 770--778.

\bibitem{Russakovsky2015ImageNet}
O.~Russakovsky, J.~Deng, H.~Su, J.~Krause, S.~Satheesh, S.~Ma, Z.~Huang,
  A.~Karpathy, A.~Khosla, M.~S. Bernstein, A.~C. Berg, and F.~Li, ``Imagenet
  large scale visual recognition challenge,'' \emph{International Journal of
  Computer Vision}, vol. 115, no.~3, pp. 211--252, 2015.

\bibitem{Wang_2020_CVPR}
X.~Wang, Y.~Liu, C.~Shen, C.~C. Ng, C.~Luo, L.~Jin, C.~S. Chan, A.~v.~d.
  Hengel, and L.~Wang, ``On the general value of evidence, and bilingual
  scene-text visual question answering,'' in \emph{IEEE/CVF Conference on
  Computer Vision and Pattern Recognition (CVPR)}, June 2020.

\end{thebibliography}

\ifCLASSOPTIONcaptionsoff
  \newpage
\fi

\end{document}